\documentclass[conference]{IEEEtran}
\makeatletter
\usepackage{cite}
\ifCLASSINFOpdf
   \usepackage[pdftex]{graphicx}
\else
\fi
\usepackage{array}
\newcolumntype{P}[1]{>{\centering\arraybackslash}p{#1}}
\newcolumntype{M}[1]{>{\centering\arraybackslash}m{#1}}
\usepackage{array}
\usepackage{url}
\begin{document}
\title{An Efficient Single Chord-based Accumulation Technique (SCA) to Detect More Reliable Corners}
\author{\IEEEauthorblockN{Mohammad Asiful Hossain,
		Abdul Kawsar Tushar, and Shofiullah Babor
	}
	\IEEEauthorblockA{Computer Science and Engineering Department, University of Asia Pacific, Dhaka, Bangladesh\\asif.hossain@uap-bd.edu, tushar.kawsar@gmail.com}
}
\maketitle
\begin{abstract}
Corner detection is a vital operation in numerous computer vision applications. The Chord-to-Point Distance Accumulation (CPDA) detector is recognized as the contour-based corner detector producing the lowest localization error while localizing corners in an image. However, in our experiment part, we demonstrate that CPDA detector often misses some potential corners. Moreover, the detection algorithm of CPDA is computationally costly. In this paper, We focus on reducing localization error as well as increasing average repeatability. The preprocessing and refinements steps of proposed process are similar to CPDA. Our experimental results will show the effectiveness and robustness of proposed process over CPDA.
\end{abstract}
\IEEEpeerreviewmaketitle

\section{Introduction}

Image processing is one of the buoyant research fields around the globe. We can define image processing as a set of operations used to extract feature information from a digital image. One of the challenges in digital image processing is to detect or match features in a digital image. A feature is a piece of relevant image information with help of which various computational tasks can be performed that are related to a certain range of related but more complex applications. Instances of these applications include edge detection, object detection, corner detection etc.  Feature detection and feature matching are The two major fields of image processing. Feature detection is a method to identify any significant property of an image like corner detection, edge detection etc. In feature matching we find similarities for test and original images. 
However, the goal of this research is to detect corners from an image by using significant corner detection technique. A corner is a salient feature that could be termed as location of an image edge where angle of the slope changes abruptly \cite{hossain2017chord}. There are two types of corner detectors - intensity based and contour based. Intensity based corner detectors directly deal with the intensity values of an image. In contrast, contour based corner detectors extract the curves from the image at the beginning, and afterward identify the locations which have salient information or maximal curvature. In this literature, we will focus on contour based corner detectors.


Recently there are a few works based on corner detectors that are chord-based. Chord-to-Point Distance Accumulation (CPDA) \cite{awrangjeb2008robust} showed outstanding performance in terms of localization error among corner detectors. This technique uses multiple chords and calculates curvature values of each point a measurement of being a corner. CPDA technique has quite low level of localization error; nonetheless, since the process uses several chord length and two separate refinement processes, it is highly expensive in terms of computational cost. As CPDA is providing the lowest localization error, we will propose a method which will produce better localization error than CPDA in the next section. At the same time, we will try increase the average repeatability with the same method. In experiment section we will see the novelty of the proposed process.

\section{Contribution}
The main contribution of this paper includes producing a better localization error than CPDA. The better the localization error is, the reliable the corner sets are. We also focus on bringing down the calculation complexity to one-third of CPDA along with producing better average repeatability. Moreover, our proposed process is detecting some potential corner points missed by CPDA. In this paper, we will demonstrate the proposed process in detail.

\section{Background}\label{background}

Maximum of the contour-based corner detectors follow the same preprocessing steps. First of all, CPDA takes RGB image as input. Then it converts the RGB image to gray scale image. With the help of canny edge detector CPDA finds out the edge image from the gray scale input image \cite{canny1986computational}. Afterwards, it extracts some particular curves from edged image and uses Gaussian Kernel to smooth those extracted curves. Next, it finds T junction points on the curve. T junction points are considered as definite corner points.  The most important and primitive part of considering a point as a corner is to calculate the curvature value of that point, and the curvature value determination process follows the preprocessing steps. After that, two basic refinement process take place in CPDA. The first one is to measure the curvature value and compare it with the threshold value to sort out the candidate corner sets. The later one excludes the false corners from the set of candidate corners to achieve the final corner set. They will be discussed shortly at the end of the section.


\begin{figure}[bh]
	\centerline{\includegraphics[width=\linewidth]{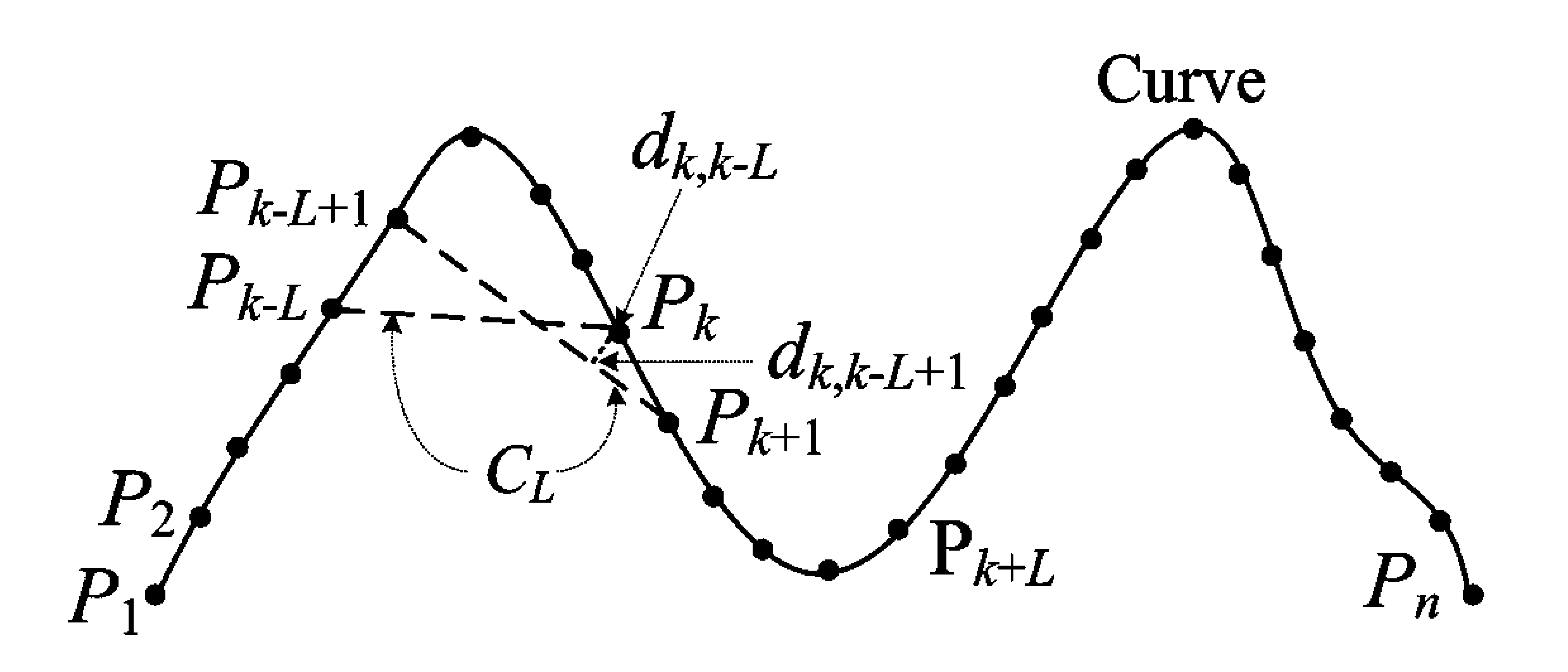}}
	\caption{Curvature value Estimation according to CPDA using chord \cite{awrangjeb2008robust}}
	\label{fig:cpda_process_good_fig}
\end{figure}

CPDA curvature value determination process is done with help of three chords of different lengths. The chord lengths are 10, 20 and 30. The chord lengths are defined to be $L_i$, where $i$ belongs to the set \{10, 20, 30\}. Finding out the value of curvature for every single point of the curve using chord length $i$ is done. In Fig. \ref{fig:cpda_process_good_fig} we show $P_1$, $P_2$, $P_3$, $...$, $P_n$ are $n$ points on the curve. From point $P_k$ using the chord $L_i$, CPDA draws a straight line to point $P_{k-L}$; then CPDA traverses with the chord to the right of the curve until $P_k$ is no longer an internal point in that curve. Therefore, at the last step of iteration, one end of the chord is at $P_k$ and the other end is at $P_{k+L}$. At each point of this iteration, distance between point $P_k$ and the straight line between two points of the chord at that step is calculated. The distance between point $P_k$ and the straight line is calculated using (\ref{eq_dist_accumulation}).

\begin{equation}\label{eq_dist_accumulation}
h_{L_i}(k)=\sum\limits_{j=k-i+1}^{k-1} {d_{k, j}}
\end{equation}

After that the curvature value for each chord length normalized because the values become in range [0,1] using (\ref{eq_normalize}).

\begin{equation}\label{eq_normalize}
	h'{_{L_i}(k)}=\frac{h_{L_i}(k)}{max(h_{L_i})}, \mbox{$i \in \{ 10, 20, 30$\}, $1\leq k \leq N$}
\end{equation}

Since there are three chords of different lengths, the entire process of curvature value determination is repeated three times, and same complex calculations involving square root operation is executed thrice. To find a candidate corner, we need a stable value from those three curvature values. Therefore, the curvature values for three chord lengths are multiplied using (\ref{eq_cpda_multiply}).

\begin{equation}\label{eq_cpda_multiply}
    H(k)=h'_{L_{10}}(k) \times h'_{L_{20}}(k) \times h'_{L_{30}}(k), \mbox{for $1\leq k \leq N$}
\end{equation}

Now CPDA aims to find out the candidate corner set from local maxima. The corner set consists of strong, weak, and false corners. The strong corners are significantly sharp and are included in the final corner set. On the other hand, the weak corners are relatively flat and therefore not considered for the final corner set. Similarly, the false corners add noise to the data set. The false and weak corners are removed by way of two refinement processes.

\begin{enumerate}[
	\setlength{\IEEElabelindent}{\dimexpr-\labelwidth-\labelsep}
	\setlength{\itemindent}{\dimexpr\labelwidth+\labelsep}
	\setlength{\listparindent}{\parindent}
	]
	\item First Refinement process:  Items of the candidate corner set are in range [0, 1] and the definition of angle threshold is set as $t_h= 0.2$ which works well. If the value of curvature of candidate corner is less than the angle threshold then that corner is declared as a weak corner and consequently removed from the candidate corner set.
    
	\item Second Refinement Process: Angle threshold is used to remove false corners. The experimental result shows that if the angle is greater than $157^\circ$ than it becomes unreasonably noisy. Therefore we define the angle threshold as $\delta=157^\circ$. When the candidate corner being evaluated ha a curvature value greater than the defined angle threshold, it is termed as a false corner and then removed.
\end{enumerate}

\section{Proposed Method}\label{proposed}

The proposed method is named as single chord-based accumulation (SCA) technique. Before detecting the corner, we need some preprocessing of the image, which are the same as those of CPDA. Applying Canny edge detector, we find the edge image from input image. Then we extract planar curves from the edge image and find T-junction and then put these T-junction in candidate corner set. We apply small scale Gaussian kernel to reduce the noise from the curves. For each curve, we estimate the curvature value at each point situated on that curve. Instead of using three chord length as is done in CPDA, we use a single chord in order to calculate curvature value. Extensive experiments were conducted with chord length values ranging from $7$ to $31$, angle values ranging from $148^\circ$ to $160^\circ$, and curvature threshold values ranging from $0.01$ to $0.2$. By observing the results of experimentation, chord length 15 is chosen as the length of chord in our process. This is because at this particular chord length, values of the two metrics of our experiment, namely - average repeatability and localization error are the most improved. More about evaluation metrics will be discussed in the subsequent section.

An intuitive reason behind choosing 15 as chord length is that, when the chord length is too small, many extraneous corners are detected which contain numerous noises and trivial details. On the other hand, when the chord length is larger, then the problem is that many important corner are missing. This is why we select a chord length that is neither too small nor too large, so as to minimize problems on both fronts.

\begin{figure}[bh]
	\centerline{\includegraphics[width=68.7mm]{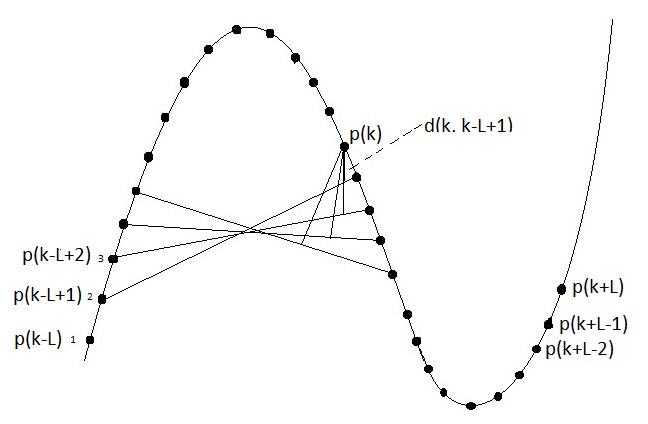}}
	\caption{Estimation of curvature value using MCPDA}
	\label{fig:mcpda_process}
\end{figure}

Our proposed method is demonstrated using Fig. \ref{fig:mcpda_process}. Let $p_1$, $p_2$, $p_3$...$p_n$ be the points of the curve. To measure the curvature value $h_{L}(k)$ at a point $p_k$ we use the chord L. When we want to measure the value of curvature at point $p_k$, we traverse with the chord from point $p_{k-L}$ to point $p_{k+L}$. We term the perpendicular distance from $p_{k-L}$ to $p_k$ as $d_{k, k-L}$, and we start our traversal with the two ends of the chord on these two points. The value of the distance is obtained by using simple Euclidean distance formula which is given in (\ref{eq:Euclidean}).

\begin{equation}\label{eq:Euclidean}
d_{k, k-L} = \sqrt{(p_{k_x} - p_{{k-L}_x})^2 + 
(p_{k_y} - p_{{k-L}_y})^2}
\end{equation}

Then we move the chord one place right in the figure when the end points are now $p_{k-L+1}$ and $p_{k+1}$ respectively. At this step, the distance value will be $d_{k, k-l+1}$ since we are still in the process of measuring the curvature value of point $p_k$. The process terminates when the two end points of the chord becomes $p_k$ and $p_{k+l}$. Finally we can estimate the discrete curvature for the point $p_k$ using (\ref{eq_dist_accumulation_mctaa}).

\begin{equation}\label{eq_dist_accumulation_mctaa}
h_{L}(k)=\sum\limits_{j=k-L}^{k} {d_{k, j}}
\end{equation}

We observe that the perpendicular distances from $p_k$ to $p_{k-l}$, as well as  from $p_k$ to $p_k$ are both zero; therefore, we optimize from (\ref{eq_dist_accumulation_mctaa}) to (\ref{eq_dist_accumulation2_mtcaa}).

\begin{equation}\label{eq_dist_accumulation2_mtcaa}
h_{L}(k)=\sum\limits_{j=k-L+1}^{k-1} {d_{k, j}}
\end{equation}

Depending on the curve, the value of $h_L(k)$ can range from zero to a large integer number. When considering this value with a threshold to remove weak or false corners, we have to normalize the value of $h_L(k)$. For each point, every distance is divided by the maximum distance thus the value is squashed within range [0, 1] following (\ref{eq_normalize_mctaa}).

\begin{equation}\label{eq_normalize_mctaa}
h'_L{(k)}=\frac{h_L(k)}{max(h_L)}, \mbox{$1\leq k \leq N$}
\end{equation}

Finally, we produce a set of candidate corner that includes weak corners, false corners, and also strong corners. Applying refinement process we remove the weak and false corners from our candidate corner set. Weak corners are flat and less significant; furthermore, false corners are noisy in nature. On the other hand, strong corner are sharper and more significant which is why they are finally added to the dataset. 

Now we follow the two step refinement process to remove the weak and false corners. These refinement processes are identical to the ones used by the CPDA process.

(1)	Using Curvature-Threshold: After normalizing the value we find a candidate corner set with value range at [0, 1]. To remove the corners that are weak, we use curvature threshold which is $T_h$. When the value of our candidate corner is less than our threshold, then we can call it as a weak corner. After conducting substantial experiments, we realize that the value of $0.067$ gives the best result to remove the weak corner. So we declare that our curvature threshold $T_h=0.067$. Using first refinement process, we are able to remove the weak corners from our candidate corner set.

(2)	Using Angle-Threshold: Using two nearest corners, we find an angle to our candidate corner which is defined by C. When this angle become larger than our threshold angle, the candidate corner is declared as a false corner. Our experimental result shows that the angle $157^\circ$ performs best than others. This means that when both the neighbors of a candidate corner is close to making a straight line with the candidate corner in middle, the candidate corner has less chance of being an actual corner. Therefore, when C becomes larger than $157^\circ$, the candidate corner is define as a false corner and subsequently removed from our candidate corner set. This false corner removal process is same as CPDA second step refinement.

\subsection{Image Accumulation and Transformation}

In our experiment we mainly use the total number of different images which totals to 23. The entire image data set is gray-scale accommodating included both artificial images as well as real world ones. This same type of image are used in CPDA technique. The greater portion of image are collected from standard database. Table \ref{table_database} elucidate the transformation which are applied to those image and become established total number of 8350 transformed test images.

\begin{table}[!h]
	\centering
	\caption{Applied several Image Transformations on 23 images from standard dataset \cite{DBLP:conf/icsipa/HossainMS15}}
	\begin{small}
		{\begin{tabular}{|M{2cm}|M{3cm}|M{2cm}|}
				\hline
				\bfseries Transformations & \bfseries Transformation factors & \bfseries Number of images \\ \hline
				Scaling & Scaling factors $s_x$=$s_y$ in [0.5, 2.0] at .1 intervals, excluding 1.0 & 345 \\ \hline
				Shearing & Shearing factors $sh_x$ and $sh_y$ in [0, 0.012]at interval=0.002. & 1081 \\ \hline
				Rotation & 18 distinct angles of range $-90^\circ$ to $+90^\circ$ at $10^\circ$  & 437 \\ \hline
				Rotation-Scale & in [-30, +30 ] at $10^\circ$ intervals, followed by uniform and non uniform factors of scale $s_x$ and $s_y$ in [0.8, 1.2] at 0.1 apart. & 4025 \\ \hline
				Nonuniform Scale & Scaling factors $s_x$ in [0.7, 1.3] and $s_y$ in [0.5, 1.5] at 0.1 intervals. & 1772 \\ \hline
				JPEG compression & Compression at 20 quality factors in [5, 100] at interval=5. & 460 \\ \hline
				Gaussian noise & Gaussian (G) noise at variance=10 in [0.005, 0.05] at interval=0.005. & 230 \\ \hline
		\end{tabular}}{\label{table_database}}
	\end{small}
\end{table}

\subsection{Evaluation Metrics}

In corner detection technique we principally use two metrics which are average repeatability and localization error. Average repeatability is used to identify the similar corners between test image and original image. Along with this, the use of localization error is to measure the error of the location between test and original images. These two techniques are used in order to gauge the performance of any corner detection technique. When a corner is found within a 3 pixel radius of original corner location, that is declare to be a repeated corner \cite{PE:PG:PG2011short:037-042, awrangjeb2008robust}. Let $A_p$ be defined as number of repeated corners, as well as $B_q$ and $C_r$ be defined as number of original and test image corners. Following these parameters, average repeatability is defined by (\ref{eval_metr}).

\begin{equation}\label{eval_metr}
Average Repeatability = 100\% \times \frac{\frac{A_p}{B_q}+
	\frac{A_p}{C_r}}{2}
\end{equation}

Localization error is measured by root means square operation using original and test image. Let the $i_{th}$ position of repeated corner in the original and test images be respectively $(x_{mi}, x_{ni})$ and $(y_{mi}, y_{ni})$. Then localization error is outlined in \ref{mscad}.

\begin{equation}\label{mscad}
L_e = \sqrt{\frac{1}{N_m}\sum_{i=1}^{N_m}(x_{oi} - x_{ti})^2 + (y_{oi} - y_{ti})^2}
\end{equation}

\subsection{Results and Discussion}

\begin{figure}[!t]
	\centerline{\includegraphics[width=68.7mm]{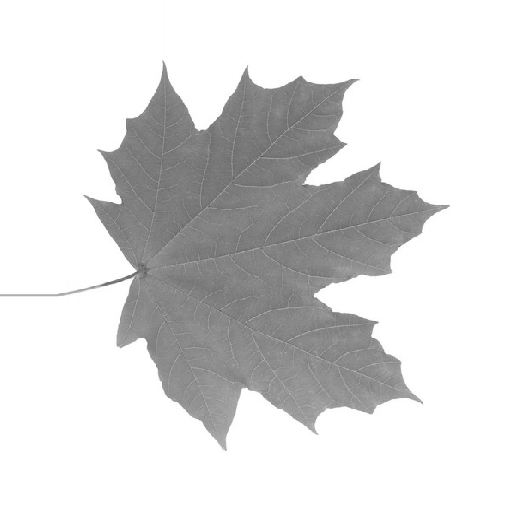}}
	\caption{Test image contained in experiment database \cite{DBLP:conf/icsipa/HossainMS15}}
	\label{fig:leaf}
\end{figure}

\begin{figure}[!t]
	\centering{\includegraphics[width=68.8mm]{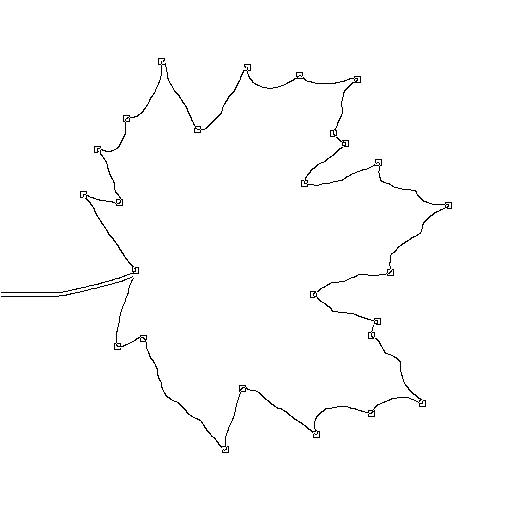}}
	\caption{Corners detected by CPDA process \cite{DBLP:conf/icsipa/HossainMS15}}
	\label{fig:CPDA}
\end{figure}

\begin{figure}[bh]
	\centerline{\includegraphics[width=68.7mm]{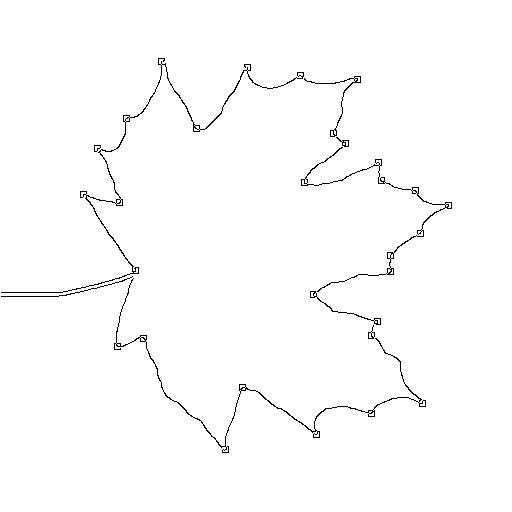}}
	\caption{Corners detected by proposed method}
	\label{fig:mcpda_leaf}
\end{figure}

In Fig. \ref{fig:leaf} we show one image of a leaf from the test image database that has been used in our experiment. Fig. \ref{fig:CPDA} depicts the corners that have been detected from the leaf image by CPDA detector, and Fig. \ref{fig:mcpda_leaf} shows the corners detected from the leaf image by the proposed detector.

As can be clearly seen from the figures of the leaves, SCA method is superior to CPDA in terms of ability of detecting corners in a figure. If we count the number of corners detected by these methods, we see that CPDA can identify 26 corners from the leaf image, while the same figure is the source of 30 detected corners from SCA detector. Among the corners that are detected by SCA but not CPDDA, some are vital and prominent corners that are clearly visible by naked eye. This demonstrates that the proposed corner detector has a superior corner detection ability in comparison to CPDA.

While an issue can be raised about the validity of the extra corners detected by our proposed method, we present Table \ref{tab:basic} as reply which demonstrates the counts of corners that have been detected from the original images and the average repeatability (across all transformations) by CPDA and SCA corner detectors. The data shows that SCA method has both higher average repeatability as well as lower localization error that CPDA detector, which are both markers of better performance.

\begin{table}[!ht]
\centering
\caption{Average repeatability and number of corners detected by corner detectors}
\label{tab:basic}
\begin{tabular}{|c|c|c|c|}
\hline
\textbf{Method} & \textbf{\begin{tabular}[c]{@{}c@{}}Average\\ Repeatability\end{tabular}} & \textbf{\begin{tabular}[c]{@{}c@{}}Localization\\ Error\end{tabular}} & \textbf{\begin{tabular}[c]{@{}c@{}}Corner\\ Count\end{tabular}} \\ \hline
CPDA            & 72.5                                                                     & 1.1437                                                                & 882                                                             \\ \hline
SCA             & 74.22                                                                    & 1.1426                                                                & 1119                                                            \\ \hline
\end{tabular}
\end{table}


\begin{figure}[!t]
	\centering{\includegraphics[width=90mm]{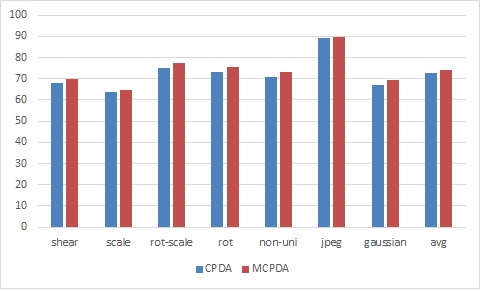}}
	\caption{Average repeatability  on various transformed images}
	\label{fig:Repeatability}
\end{figure}

\begin{figure}[!t]
	\centering{\includegraphics[width=90mm]{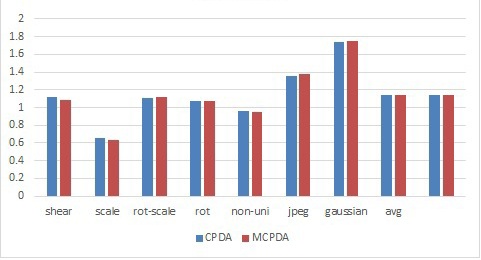}}
	\caption{Localization of error of corner detectors on various transformed images}
	\label{fig:Localization}
\end{figure}

The reason for this superiority of SCA method over CPDA is in part attributed to the reduced complexity in calculation. Whereas CPDA uses three chords of different lengths, SCA uses only a single chord of unique length to do the same calculations; as a result the number of calculations is reduced to one-third of that of CPDA. Among all the calculations done\newpage
to estimate curvature value, the square root operation is the most computationally expensive one \cite{PE:PG:PG2011short:037-042, teng2015effective}. It is so expensive that other common arithmetic operations performed in the corner detection process are relatively insignificant. Therefore, in our method, when we reduce the number of square root operations to one-third of that of CPDA, it really brings a performance boost for SCA detector operations.

\section{Conclusion}

In this paper, we have proposed a modified and effective contour based corner detector. In comparison to CPDA detector, the proposed detector achieves lower localization error with better average repeatability that can result in detecting reliable corners. The proposed SCA detector is able to detect larger number of corners than CPDA. In addition, our proposed detector is also computationally much faster than the CPDA detector.

\bibliography{bib}{}

\begin{thebibliography}{1}
\providecommand{\url}[1]{#1}
\csname url@samestyle\endcsname
\providecommand{\newblock}{\relax}
\providecommand{\bibinfo}[2]{#2}
\providecommand{\BIBentrySTDinterwordspacing}{\spaceskip=0pt\relax}
\providecommand{\BIBentryALTinterwordstretchfactor}{4}
\providecommand{\BIBentryALTinterwordspacing}{\spaceskip=\fontdimen2\font plus
\BIBentryALTinterwordstretchfactor\fontdimen3\font minus
  \fontdimen4\font\relax}
\providecommand{\BIBforeignlanguage}[2]{{%
\expandafter\ifx\csname l@#1\endcsname\relax
\typeout{** WARNING: IEEEtran.bst: No hyphenation pattern has been}%
\typeout{** loaded for the language `#1'. Using the pattern for}%
\typeout{** the default language instead.}%
\else
\language=\csname l@#1\endcsname
\fi
#2}}
\providecommand{\BIBdecl}{\relax}
\BIBdecl

\bibitem{hossain2017chord}
M.~A. Hossain and A.~K. Tushar, ``Chord angle deviation using tangent (cadt),
  an efficient and robust contour-based corner detector,'' in \emph{Imaging,
  Vision \& Pattern Recognition (icIVPR), 2017 IEEE International Conference
  on}.\hskip 1em plus 0.5em minus 0.4em\relax IEEE, 2017, pp. 1--6.

\bibitem{awrangjeb2008robust}
\BIBentryALTinterwordspacing
M.~Awrangjeb and G.~Lu, ``Robust image corner detection based on the
  chord-to-point distance accumulation technique,'' \emph{{IEEE} Trans.
  Multimedia}, vol.~10, no.~6, pp. 1059--1072, 2008. [Online]. Available:
  \url{http://dx.doi.org/10.1109/TMM.2008.2001384}
\BIBentrySTDinterwordspacing

\bibitem{canny1986computational}
J.~Canny, ``A computational approach to edge detection,'' \emph{Pattern
  Analysis and Machine Intelligence, IEEE Transactions on}, no.~6, pp.
  679--698, 1986.

\bibitem{DBLP:conf/icsipa/HossainMS15}
\BIBentryALTinterwordspacing
M.~A. Hossain, N.~Mohammed, and R.~M.~N. Sadat, ``Chord to triangular arm angle
  ({CTAA}), a more accurate version of the {CTAR} corner detector,'' in
  \emph{2015 {IEEE} International Conference on Signal and Image Processing
  Applications, {ICSIPA} 2015, Kuala Lumpur, Malaysia, October 19-21, 2015},
  2015, pp. 203--207. [Online]. Available:
  \url{http://dx.doi.org/10.1109/ICSIPA.2015.7412190}
\BIBentrySTDinterwordspacing

\bibitem{PE:PG:PG2011short:037-042}
R.~M.~N. Sadat, S.~W. Teng, and G.~Lu, ``{An Effective and Efficient
  Contour-based Corner Detector using Simple Triangular Theory},'' in
  \emph{Pacific Graphics Short Papers}, B.-Y. Chen, J.~Kautz, T.-Y. Lee, and
  M.~C. Lin, Eds.\hskip 1em plus 0.5em minus 0.4em\relax The Eurographics
  Association, 2011.

\bibitem{teng2015effective}
S.~W. Teng, R.~M.~N. Sadat, and G.~Lu, ``Effective and efficient contour-based
  corner detectors,'' \emph{Pattern Recognition}, vol.~48, no.~7, pp.
  2185--2197, 2015.

\end{thebibliography}
\bibliographystyle{IEEEtran}
\end{document}